\documentclass[10pt,twocolumn,letterpaper]{article}
\usepackage{iccv}
\pdfoutput=1
\usepackage{times}
\usepackage{epsfig}
\usepackage{graphicx}
\usepackage{amsmath}
\usepackage{amssymb}

\usepackage{authblk}
\newcommand*\samethanks[1][\value{footnote}]{\footnotemark[#1]}

\usepackage{booktabs}
\usepackage{tabulary}
\usepackage{ragged2e}
\usepackage{array}

\newcolumntype{L}[1]{>{\raggedright\let\newline\\\arraybackslash\hspace{0pt}}m{#1}}
\newcolumntype{C}[1]{>{\centering\let\newline\\\arraybackslash\hspace{0pt}}m{#1}}
\newcolumntype{R}[1]{>{\raggedleft\let\newline\\\arraybackslash\hspace{0pt}}m{#1}}

\usepackage[normalem]{ulem}
\usepackage{enumitem}
\usepackage[footnotesize]{subfigure}

\iccvfinalcopy 

\def\iccvPaperID{1037} 

\newcommand\blfootnote[1]{%
  \begingroup
  \renewcommand\thefootnote{}\footnote{#1}%
  \addtocounter{footnote}{-1}%
  \endgroup
}

\begin{document}

\title{Group Re-Identification via \\Unsupervised Transfer of Sparse Features Encoding}

\author[1]{Giuseppe Lisanti\thanks{G. Lisanti and N. Martinel should be considered as joint first-authors.}$^,$}
\author[2]{Niki Martinel\samethanks$^,$}
\author[1]{Alberto Del Bimbo}
\author[2]{Gian Luca Foresti}
\affil[1]{MICC - University of Firenze, Italy}
\affil[2]{AViReS Lab - University of Udine, Italy}
\affil[ ]{\tt\small giuseppe.lisanti@unifi.it, niki.martinel@uniud.it}
\affil[ ]{\tt\small alberto.delbimbo@unifi.it, gianluca.foresti@uniud.it}
\renewcommand\Authands{ and }

\maketitle

\begin{abstract}
Person re-identification is best known as the problem of associating a single person that is observed from one or more disjoint cameras.
The existing literature has mainly addressed such an issue, neglecting the fact that people usually move in groups, like in crowded scenarios.
We believe that the additional information carried by neighboring individuals provides a relevant visual context that can be exploited to obtain a more robust match of single persons within the group.
Despite this, re-identifying groups of people compound the common single person re-identification problems by introducing changes in the relative position of persons within the group and severe self-occlusions.
In this paper, we propose a solution for group re-identification that grounds on transferring knowledge from single person re-identification to group re-identification by exploiting sparse dictionary learning.
First, a dictionary of sparse atoms is learned using patches extracted from single person images.
Then, the learned dictionary is exploited to obtain a sparsity-driven residual group representation, which is finally matched to perform the re-identification.
Extensive experiments on the i-LIDS groups and two newly collected datasets show that the proposed solution outperforms state-of-the-art approaches.\blfootnote{This research was partially supported by the Social Museum and Smart Tourism, MIUR project no. CTN01\_00034\_23154\_SMST and by the ``PREscriptive Situational awareness for cooperative autoorganizing aerial sensor NETworks'' project CIG68827500FB.}
\end{abstract}

\section{Introduction}
\label{sec:introduction}
Person re-identification is the problem of associating a \textit{single} person that moves across disjoint camera views.
The open challenges like changes in viewing angle, background clutter, and occlusions have recently yield to a surge of effort by the community~\cite{Vezzani2014}.
In particular, existing works have focused on seeking either the best feature representations (\eg,~\cite{Lisanti2014,Matsukawa2016,Liao2015}) or propose to learn optimal matching metrics (\eg,~\cite{Kostinger2012,Liao2015a,Zhang2016}).
Despite obtaining interesting results on benchmark datasets (\eg,~\cite{Garcia2016,Paisitkriangkrai2015,Zheng2016}, such works have generally neglected the fact that in crowded public environments people often walk in \textit{groups}.

\begin{figure}[t]
\centering
\subfigure[]{\label{fig:switch}\includegraphics[width=0.17\textwidth]{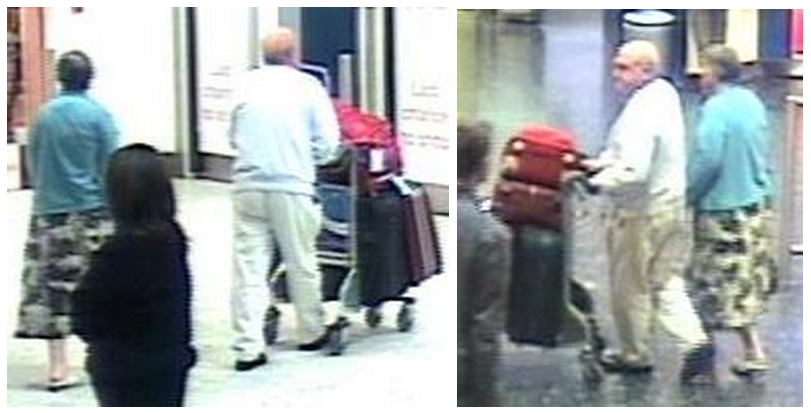}}\hfill
\subfigure[]{\label{fig:backg}\includegraphics[width=0.127\textwidth,trim={0 0 23cm 0},clip]{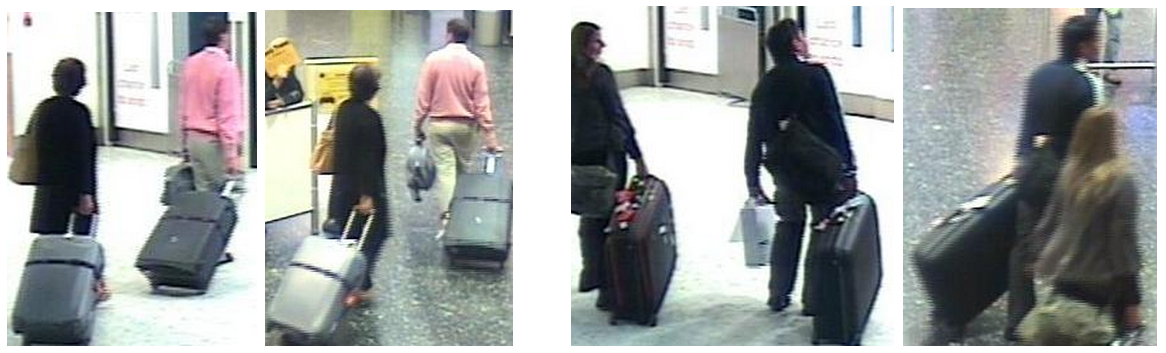}}\hfill
\subfigure[]{\label{fig:occl}\includegraphics[width=0.15\textwidth,trim={20cm 0 0 0},clip]{group2.png}}
\caption{Major group re-identification issues. Examples of: \subref{fig:switch} position swap between person in a group; \subref{fig:backg} different background between group images; \subref{fig:occl} group images with partial occlusion.}
\label{fig:group2}
\end{figure}

We believe that being able to associate the same \textit{group of people} can be a powerful tool to improve classic \textit{single-person} re-identification.
Indeed, the appearance of the whole group provides a rich visual context that can be extremely useful to reduce the ambiguity in retrieving those persons that are partially occluded or to understand the behavior of the group over time if a person in the group is missed for a certain period of time.

Group re-identification introduce some additional difficulties with respect to classic person re-identification (see~\figurename~\ref{fig:group2}).
First of all, the focus is no longer on a single subject, hence the visual appearance of all the persons in the group should be considered.
The relative displacement of the subjects in a group can be different from camera to camera.
Self-occlusions or occlusions generated by other people near by, as well as the fact that an individual in a group may be missing because he/she left the group, bring in additional challenges.
Such challenges deny the direct application of existing representation descriptors and matching methods for single-person re-identification to the group association problem.
\textit{In this work, we investigate the problem of associating groups of people}.

\noindent\textbf{Contribution:}
The contribution of this work is twofold:
i) To handle the spatial displacement configuration of persons in a group, we introduce a visual descriptor that is invariant both to the number of subjects and to their displacement within the image. 
Such a task is accomplished by 
ii) introducing a sparse feature encoding solution that leverages on the knowledge that can be acquired from the large quantity of data that is available in the \textit{single-person} re-identification domain and \emph{transfer} it to the \textit{group} re-identification domain in an unsupervised fashion.

To validate the proposed solution, we compare with existing methods on the i-LIDS group benchmark dataset.
In addition, to study the behavior of the approach under different conditions, we have collected two new group re-identification datasets.
Extensive evaluations demonstrate that better performances than current solutions are obtained on all datasets.

\section{Related Work}
While being a young field of research, the community has recently produced several works to address the re-identification problem~\cite{Vezzani2014}.
In the following we provide a brief overview of the most relevant works to our approach.

\noindent\textbf{Single Person Re-Identification:}
The literature on single person re-identification can be clustered into two main categories:
i) direct matching and
ii) metric learning-based methods.
Works belonging to the first group aim to address the re-identification problem by designing --or learning-- the most discriminative appearance feature descriptors.
Multiple local and global feature~\cite{Bazzani2012,Bak2012,Ma2014} were combined with reference sets~\cite{An2013a}, patch matching strategies~\cite{Zhao2013}, saliency learning~\cite{Zhao2013a,Wang2014c,Martinel2015c}, joint attributes~\cite{Roth2014,Khamis2014,Layne2014} and camera network-oriented schemes~\cite{Martinel2016a}. 
Among all the methods in this category, to date, the most widely used appearance descriptors are the Gaussian of Gaussian (GOG)~\cite{Matsukawa2016}, the Local Maximal Occurrence (LOMO)~\cite{Liao2015} and the Weighted Histogram of Overlapping Stripes (WHOS)~\cite{Lisanti2014,karaman2014leveraging}.

Approaches grouped in the second family represent the trend in person re-identification.
In particular, metric learning approaches have been proposed by relaxing~\cite{Hirzer2012a} or enforcing~\cite{Liao2015a} positive semi-definite (PSD) conditions, by considering equivalence constraints~\cite{Kostinger2012,Tao2013,Tao2014} or by exploiting the null-space~\cite{Zhang2016}.
While most of the existing methods capture the global structure of the dissimilarity space, local solutions~\cite{Li2013b,Pedagadi2013,Garcia2016} have been proposed too.
Sample-specific metrics were also investigated in~\cite{Zhang2016a}.
Following the success of both approaches, methods combining them in ensembles~\cite{Paisitkriangkrai2015,Xiong2014,Martinel2016} have been introduced.
Different solutions yielding similarity measures have also been investigated by proposing to learn listwise~\cite{Chen2015a} and pairwise~\cite{Zheng2012a} similarities.

To deal with the re-identification of a single person all such works assume that the provided images represent good detections of a single person only. This limits their application when more than a person appears in the given image.

\noindent\textbf{Group Person Re-Identification:}
The first work concerning group association over space and time was proposed in~\cite{Zheng2009a}.
The authors introduced a group representation and matching algorithm based on a learned dictionary.
Since then, the literature on this task is limited to two works~\cite{Cai2010,Ukita2016}.
Specifically, in~\cite{Cai2010}, independence of persons locations within the group was captured by the covariance descriptor, while in~\cite{Ukita2016}, spatio-temporal group features were explored to improve single person re-identification.

Differently from our work, such approaches either assume that background/foreground segmentation masks are available or exploit training data coming from the same domain (\ie, dataset) of the evaluation data.

Other works have addressed the problem of group-based verification~\cite{Zheng2016} and group membership prediction~\cite{Zhang2015b}, but both tasks still assume that the input datum represents a single person only.
Group information was also explored to address visual tracking~\cite{Xu2016,Insafutdinov2016,Assari2016} and behavior analysis~\cite{Alameda-Pineda2016} among other tasks.

\noindent\textbf{Object Displacement Invariant Descriptors:}
The most relevant problem in group re-identification is determined by the fact that people often change their positions while walking in a group.
A standard approach to deal with a similar problem in image retrieval is to extract a set of local descriptors, encode them and pool them into an image-level signature which is independent from the spatial location.

\begin{figure*}[t!]
\centering
\includegraphics[width=1\textwidth]{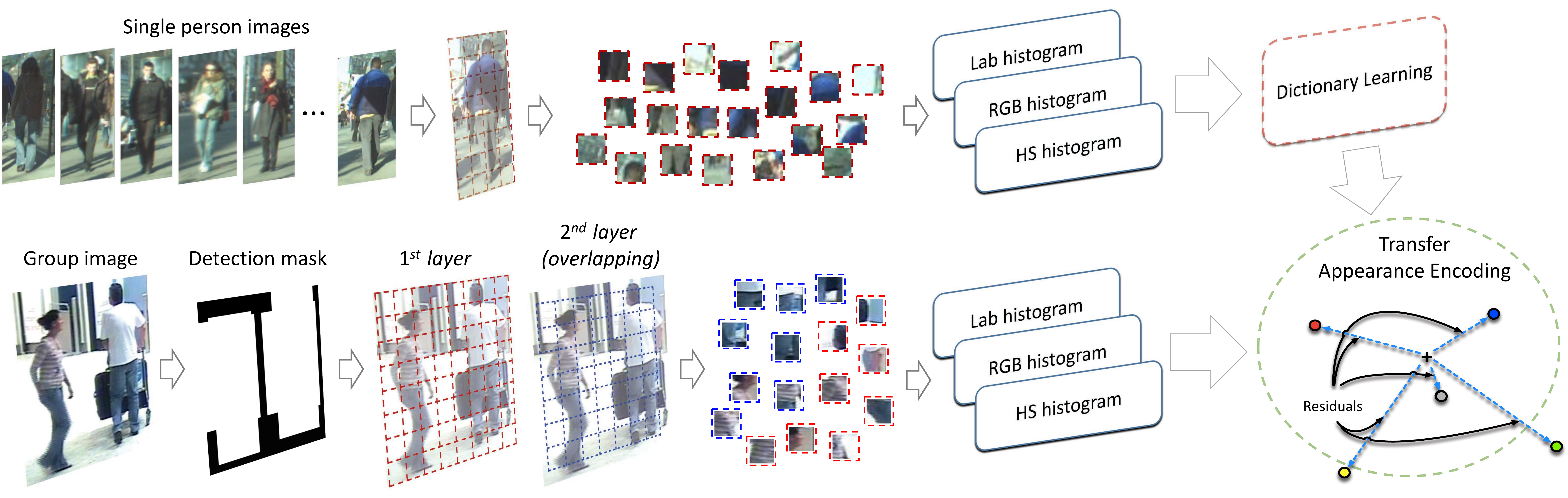}
\caption{Proposed group re-identification pipeline.
Top row shows the unsupervised single-person dictionary learning.
Bottom row depicts the re-identification process with feature extraction and subsequent sparse residual encoding obtained with the transferred dictionary atoms.
}
\label{fig:overview}
\end{figure*}

Research in this area is quite vast, but almost all approaches inherit or extend the Bag-of-Words (BoW)~\cite{Csurka2004}, the Vector of Locally Aggregated Descriptors (VLAD)~\cite{Jegou2010b} or the Fisher Vector (FV)~\cite{Sanchez2013}.
In person re-identification such solutions have been explored to encode first and second-order derivatives for each pixel~\cite{Ma2012a} and to address large-scale applications~\cite{Zheng2015d}.
Similar solutions exploiting an encoding scheme based on dictionary learning have also been proposed in~\cite{Kodirov2015,Peng_2016_CVPR}.

These works did not address the group re-identification problem and mainly adopted the encoding schemes to deal with extremely high dimensional image descriptors.

The closest work to our approach~\cite{Zheng2009a} exploited a classical BoW scheme on densely extracted features and combined them with a proposed global descriptor.
In addition, authors assumed that foreground/background segmentation masks were available such that only features extracted for foreground pixels were used to construct visual words for group image representation.
Our approach has three key differences with such a work:
i) we propose a novel encoding scheme based on dictionary learning;
ii) there is no requirement of foreground/background segmentation masks which demands a substantial hand-work;
iii) knowledge obtained from single person re-identification domain is transferred to tackle the group re-identification problem.

\section{Proposed Approach}

In the following we first define how group appearance is modeled.
Then we introduce a transfer learning solution that allows us to exploit knowledge available in single-person re-identification to better tackle the group re-identification. Finally, we describe how groups matching is performed.
The whole process is depicted in~\figurename~\ref{fig:overview}.

\subsection{Group Appearance Modeling}
\label{sec:group_features}
In our representation, the image of a group is resized to $128\times128$ pixels. Two set of patches with fixed dimension $16\times16$ are then extracted. The first set is obtained from the whole image, whereas the second one is chosen so as to collect information that overlaps with the first layer, see figure~\ref{fig:overview}.
For each patch, we compute three histograms considering the same image projected onto different color spaces, namely: HS, RGB and Lab.
For the HS images we consider 8 bins for each channel, while for the RGB and Lab we use 4 bins for each channel.
This results in a 64 dimensional histogram for each patch and color space (\eg,  $8\times8$ for HS or $4\times4\times4$ for RGB and Lab).

To obtain a representation that does not preserve location information and is more robust to changes in the group configuration, we separately consider each histogram extracted from each patch (\ie, we do not concatenate them).

Due to the unconstrained patch image subdivision, noisy background information is captured by the feature representation.
To circumvent such an issue, we first run three different person detectors, namely Deformable Part Models~\cite{dpm2010}, Aggregated Channel Feature~\cite{acfdollar} and R-CNN~\cite{girshick14CVPR}.
Then, the filtering mask obtained as the combination of the responses of these three detectors is used to weight the contribution of each pixel in the histogram computation (\ie, pixels belonging to the background have zero contribution).

\begin{figure*}
\centering
\includegraphics[width=\linewidth]{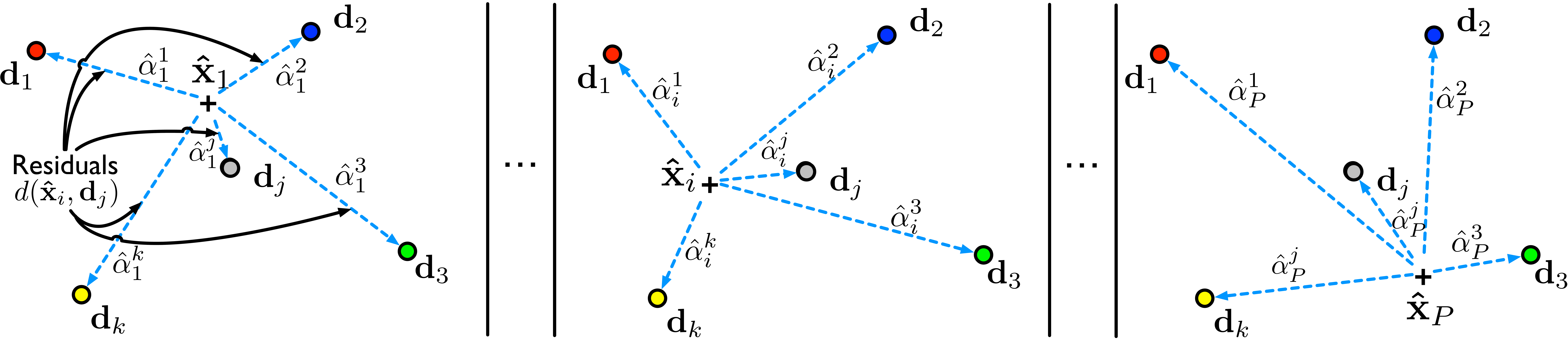}
\caption{Proposed sparse residual encoding.
Colored circles represent the learned dictionary atoms.
Black crosses denote the visual features extracted from each of the $\hat{P}$ patches into which the group image is divided.
Blue dashed arrows show the residual computed via $d(\cdot,\cdot)$, which are then weighted by the corresponding sparse reconstruction coefficients $\hat{\alpha}$.}
\label{fig:residual_encoding}
\end{figure*}

\vspace{-1em}
\subsection{Unsupervised Learning of Person Appearance}
\label{sec:unsup_dict_learning}
We propose to exploit a sparse dictionary learning framework that allows us to represent a group of persons as a combination of few human body parts (\ie, patches/atoms).
Since these atoms does not necessarily need to be structured accordingly to the relative person displacements, we obtain a flexible group representation.
Such a solution resembles visual encoding schemes (\eg, BoW~\cite{Csurka2004}, FV~\cite{Sanchez2013}, VLAD~\cite{Jegou2010b}) that are widely adopted for image classification with local descriptors.

We first exploit the dictionary learning solution in~\cite{Mairal2009} to find the basis set of patches that yields to the optimal reconstruction accuracy for single person re-identification.
Then, we leverage on such a basis set to introduce a sparse residual group representation.

\vspace{1pt}
\noindent\textbf{Problem Definition:}
Let $\mathcal{I}^{tr} = \{\mathbf{I}_1, \ldots, \mathbf{I}_N\}$ be a training set composed of $N$ images belonging to a \textit{single person} re-identification domain (\ie, images in $\mathcal{I}^{tr}$ may come from the ETHZ~\cite{Schwartz2009a}, CAVIAR~\cite{CAVIAR2004}, or VIPeR~\cite{Gray2007a} dataset).
Also let $P$ denote the number of patches into which each image is divided such that $\mathcal{X}^{\mathrm{tr}} = \{\mathbf{x}_1, \ldots, \mathbf{x}_{NP}\}$ is a training set containing $NP$ $d$-dimensional vectors $\mathbf{x}$, each representing the visual features extracted from a single patch\footnote{In our current solution, $\mathbf{x}$ represents a $64$-D histogram extracted either from the HS, RGB or Lab color space.
These are obtained with a similar approach to the one described in Sec.~\ref{sec:group_features} but with input images not processed by the detectors and resized to $128\times 64$.
}.

With this, we define our optimization objective as
\begin{equation}
\label{eq:dict_objective}
\mathcal{L}(\mathbf{D}) = \ \frac{1}{NP} \sum_{i=1}^{NP} l(\mathbf{x}_i, \mathbf{D})
\end{equation}
where $\mathbf{D} = [\mathbf{d}_1^T,\ldots,\mathbf{d}_k^T]$, with $\mathbf{d} \in \mathbb{R}^d$ is the dictionary of $k$ atoms to be learned and $l(\cdot,\cdot)$ is a suitable loss function such that its output should be ``small'' if $\mathbf{D}$ is able to provide a good representation for any training input datum $\mathbf{x}_i$.

It has been demonstrated in many fields, ranging from image compression to person re-identification itself~\cite{Karanam2016}, that obtaining a representation of a signal $\mathbf{x}$ using only a few elements of a dictionary $\mathbf{D}$ performs better than considering all the atoms.
We let our loss function $l$ be the optimal value of the $\ell_1$-sparse coding problem, \ie
\begin{equation}
\label{eq:sparse_coding}
l(\mathbf{x},\mathbf{D}) = \min_{\pmb{\alpha}} \frac{1}{2} \|\mathbf{x} - \mathbf{D}\pmb{\alpha} \|_2^2 + \lambda \|\pmb{\alpha}\|_1
\end{equation}
where $\pmb{\alpha} \in \mathbb{R}^{k}$ is the sparse vector of coefficients and $\lambda$ is a regularization parameter that balances the trade-off between a perfect reconstruction and the sparsity of $\pmb{\alpha}$.

By solving the minimization problem in eq.~(\ref{eq:sparse_coding}), we find the set of atoms in $\mathbf{D}$ that yields the best reconstruction for the signal $\mathbf{x}$.
Despite this being compliant to our objective, it does not answer the problem of finding the set of all $k$ atoms that minimize eq.~(\ref{eq:dict_objective}).

To address such a problem the $\ell_1$-sparse coding problem can be rewritten as a joint optimization whose solution should result in the best combination of dictionary atoms and sparse coefficients.
Thus, eq.~(\ref{eq:sparse_coding}) can be rewritten as 
\begin{equation}
\label{eq:sparse_coding_joint}
l(\mathbf{x}_i,\mathbf{D}) = \min_{\mathbf{D} \in \mathcal{C},\pmb{\Theta}} \frac{1}{NP} \sum_{i=1}^{NP} \left( \frac{1}{2} \|\mathbf{x}_i - \mathbf{D}\pmb{\Theta}_i \|_2^2 + \lambda \|\pmb{\Theta}_i\|_1 \right).
\end{equation}
where $\mathbf{\Theta} = [\pmb{\alpha}_1, \ldots, \pmb{\alpha}_{NP}]$ contains the sparse coefficients to be found for each patch and $\mathcal{C} = \{\mathbf{D} \in \mathbb{R}^{d\times k} | \mathbf{d}_j^T \mathbf{d}_j \leq 1, \ \forall j=1,\ldots,k \}$ is a convex set introducing an $\ell_2$ norm constraint on the single atoms.

\vspace{1pt}
\noindent\textbf{Optimization Solution:}
The problem in eq.~(\ref{eq:sparse_coding_joint}) is not jointly convex, but convex with respect to each of the two variables when the other one is fixed.
To solve the optimization problem, the solution proposed in~\cite{Mairal2009} is exploited.
It alternatively solves the classical sparse coding first, then it updates the learned dictionary using the so computed optimal sparse coefficient.

Specifically, let $t$ denote the optimization iteration counter.
Also let $\mathbf{D}_{t}$ be a randomly initialized dictionary, and $\mathbf{A}_{t} \in \mathbb{R}^{k\times k} = \mathbf{0}$ and $\mathbf{B}_t \in \mathbb{R}^{k\times k} = \mathbf{0}$ (with $t=0$) be two matrices which will carry the information of all the sparse coefficients $\pmb{\alpha}$'s.
Then, we start the optimization by randomly drawing an image training sample from the training set and computing the visual representation of a randomly chosen patch $\mathbf{x}_{t}$.
Such a datum is then considered by the \emph{least angle regression} (LARS)~\cite{efron2004least} to solve the sparse coding problem in eq.~(\ref{eq:sparse_coding}), hence to obtain the vector of sparse coefficients for the $t$-th iteration $\pmb{\alpha}_t$.

The computed sparse coefficient vector is then exploited to revise $\mathbf{A}_t$ and $\mathbf{B}_t$ such that these can be used in a block-coordinate descent solution to update the learned dictionary.
More precisely, the two matrices carry all the information brought in by all the sparse coefficients computed so far as
\begin{gather}
\mathbf{A}_{t} = \mathbf{A}_{t-1} + \pmb{\alpha}_t \pmb{\alpha}_t^T
\\
\mathbf{B}_{t} = \mathbf{B}_{t-1} + \mathbf{x}_t \pmb{\alpha}_t^T .
\end{gather}

Exploiting the block coordinate descent to update the dictionary $\mathbf{D}_t$ yields to the following solution for each dictionary atom,~\ie for each column $\mathbf{d}_j$ with $j=1,\ldots,k$
\begin{align}
\mathbf{v} & = \frac{1}{\mathrm{Tr}(\mathbf{A}_t)_j} (\mathbf{b}_j - \mathbf{D}_t \mathbf{a}_j) + \mathbf{d}_j
\\ 
\mathbf{d}_j & = \frac{1}{\max(\|\mathbf{v}\|_2, 1) } \mathbf{v}
\end{align}
where $\mathrm{Tr}(\mathbf{A}_t)_j$ is the $j$-th element on the diagonal of $\mathbf{A}_t$, while $\mathbf{a}_j$ and $\mathbf{b}_j$ are the $j$-th columns of $\mathbf{A}_t$ and $\mathbf{B}_{t}$, respectively.

The optimization is run for $T$ iterations.
Once such a limit is reached, we let $\mathbf{D}^{*} = \mathbf{D}_{T}$ be the solution for eq.~(\ref{eq:dict_objective}).


\subsection{Transfer Single-to-Group Appearance}
\label{sec:residual_encoding}
Inspired by the recent success of residual learning both for visual encoding~\cite{Jegou2010b} and for deep learning~\cite{He2016,He2016a}, we propose to exploit the single-person learned dictionary and introduce a sparsity-driven residual representation for an unseen \textit{group} image $\mathbf{\hat{I}}$.
The process is shown in~\figurename~\ref{fig:residual_encoding}.

We start by extracting the visual features from each of the $\hat{P}$ patches as computed in~Sec.~\ref{sec:group_features}.
Then, for each $\mathbf{\hat{x}}_i$ with $i=1,\ldots,\hat{P}$ we compute its \textit{residual} $d(\mathbf{\hat{x}}_i, \mathbf{d}_j)$ with every $j=1,\cdots,k$ atom in the learned dictionary $\mathbf{D}^*$.
Notice that the residual $d(\cdot,\cdot)$ can be any suitable function that describes how much ``dissimilar'' the two inputs are (\eg, the euclidean distance,~\etc).

Then, we solve the $\ell_1$-sparse coding problem in eq.~(\ref{eq:sparse_coding}) to obtain the sparse vector of coefficients $\pmb{\hat{\alpha}}_i$ for each $\mathbf{\hat{x}}_i$.
Since each patch is considered separately, every element in  $\pmb{\hat{\alpha}}_i = [\hat{\alpha}^{1}_{i},\ldots,\hat{\alpha}^{k}_{i}]$ specifies \textit{how  important} a particular atom is in the reconstruction of $\mathbf{\hat{x}}_i$.

Armed with the aforementioned results, we want to assign more importance to the residuals computed with respect to those dictionary atoms that are relevant for the sparse reconstruction of the considered sample $\mathbf{\hat{x}}_i$.
A reasonable approach to meet this objective is to weight the residuals through the corresponding sparse dictionary coefficients. This results in:
\begin{equation}
\label{eq:weighted_residuals}
\mathbf{\hat{x}}^*_i = [\hat{\alpha}_1 d(\mathbf{\hat{x}}_i,\mathbf{d}_1), \ldots, \hat{\alpha}_k d(\mathbf{\hat{x}}_i,\mathbf{d}_k)].
\end{equation}

\subsubsection{Sparse Residual Pooling}
The proposed residual representation is obtained for each of the $\hat{P}$ patches of a group.
To compute the final representation that can be used to match two groups of persons we should introduce a suitable combination of all the $\mathbf{\hat{x}}^*_i$'s.
A classical approach would be to concatenate all such elements.
However, in doing so we may lose one of the relevant features of visual encoding schemes,~\ie, represent any number of feature vectors as a sample in a feature space of fixed dimensionality.
In addition, such a solution is likely to bring in the problem of the curse of dimensionality since the final dimension is linear with respect to both $k$ and $N\hat{P}$.

To overcome these issues, we propose to use different pooling schemes that produce a compact representation, denoted $\mathbf{\hat{f}}$, that depends only on the number of atoms $k$.
Specifically, we exploited the
\textit{average pooling} (\ie, $\mathbf{\hat{f}}_j = \frac{1}{\hat{P}}\sum_{i}^{\hat{P}} \mathbf{\hat{x}}_{i,j}^*$)
and \textit{max pooling} (\ie, $\mathbf{\hat{f}}_j = \max(\mathbf{\hat{x}}_{1,j}^*, \cdots, \mathbf{\hat{x}}_{\hat{P},j}^*)$),
where 
$j=1,\ldots,k$ indicates the $j$-th element of the corresponding vectors.

\subsubsection{Group Representation and Matching}
The final group representation is computed as $\mathbf{\hat{s}} = \Phi(\mathbf{\hat{f}})$ 
where $\Phi(\cdot)$ is the Principal Component Analysis (PCA) mapping function $\mathbb{R}^{k} 
\mapsto \mathbb{R}^u$ with $u \ll k$.
With such a representation, the dissimilarity between two group images $\mathbf{\hat{I}}_A$ and $\mathbf{\hat{I}}_B$ is computed as $\delta(\mathbf{\hat{I}}_A, \mathbf{\hat{I}}_B) = \prod_f \Psi(\mathbf{\hat{s}}_A^{f}, \mathbf{\hat{s}}_B^{f})$ with $\Psi$ denoting the cosine distance and $f \in \{\mathrm{HS}, \mathrm{RGB}, \mathrm{Lab}\}$.

\begin{figure*}[t]
	\centering
    	\subfigure[i-LIDS Groups]{\label{fig:ilids_samples}\includegraphics[width=0.325\linewidth]{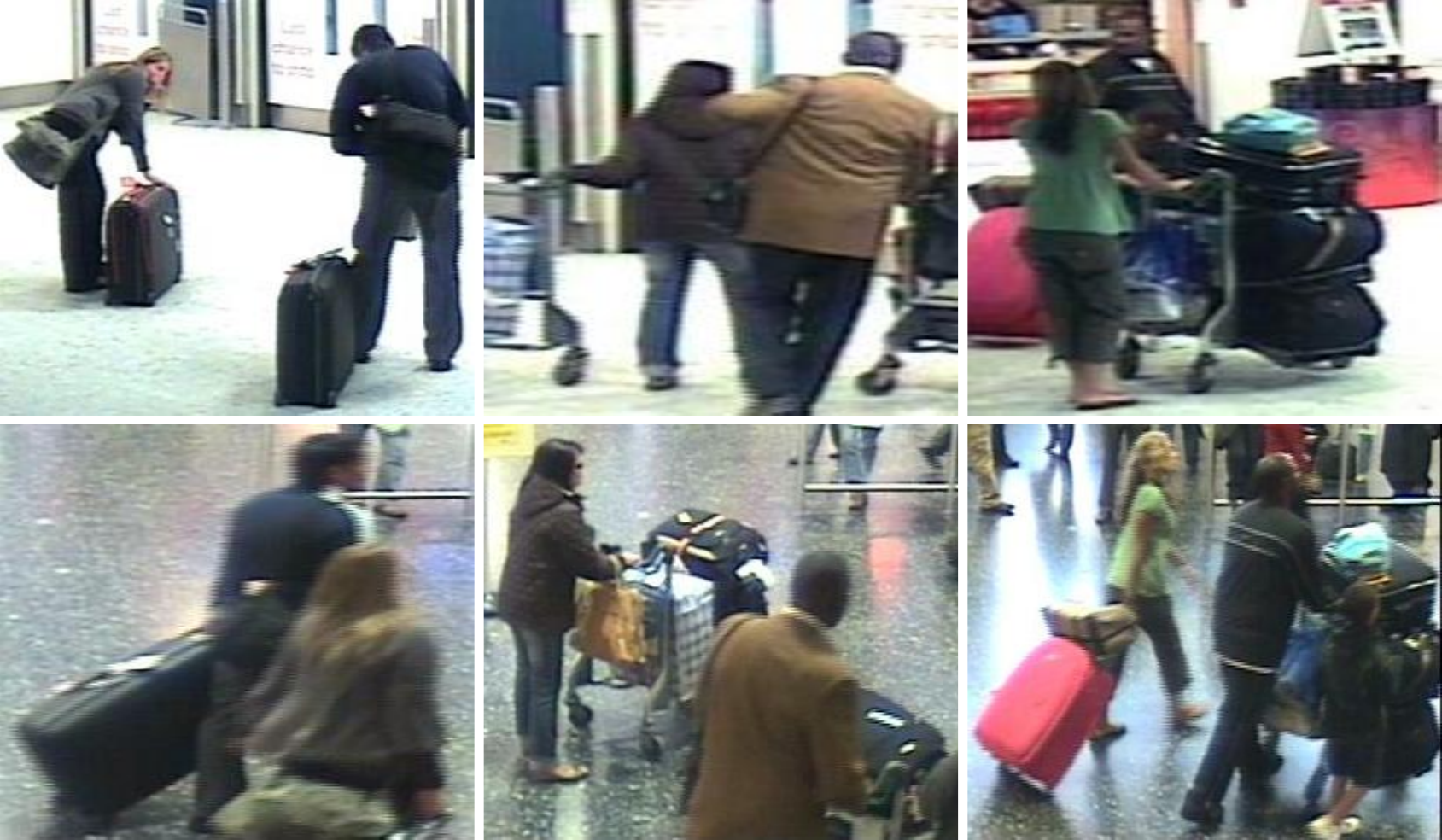}} \hfill
		\subfigure[Museum Groups]{\label{fig:museum_samples}\includegraphics[width=0.325\linewidth]{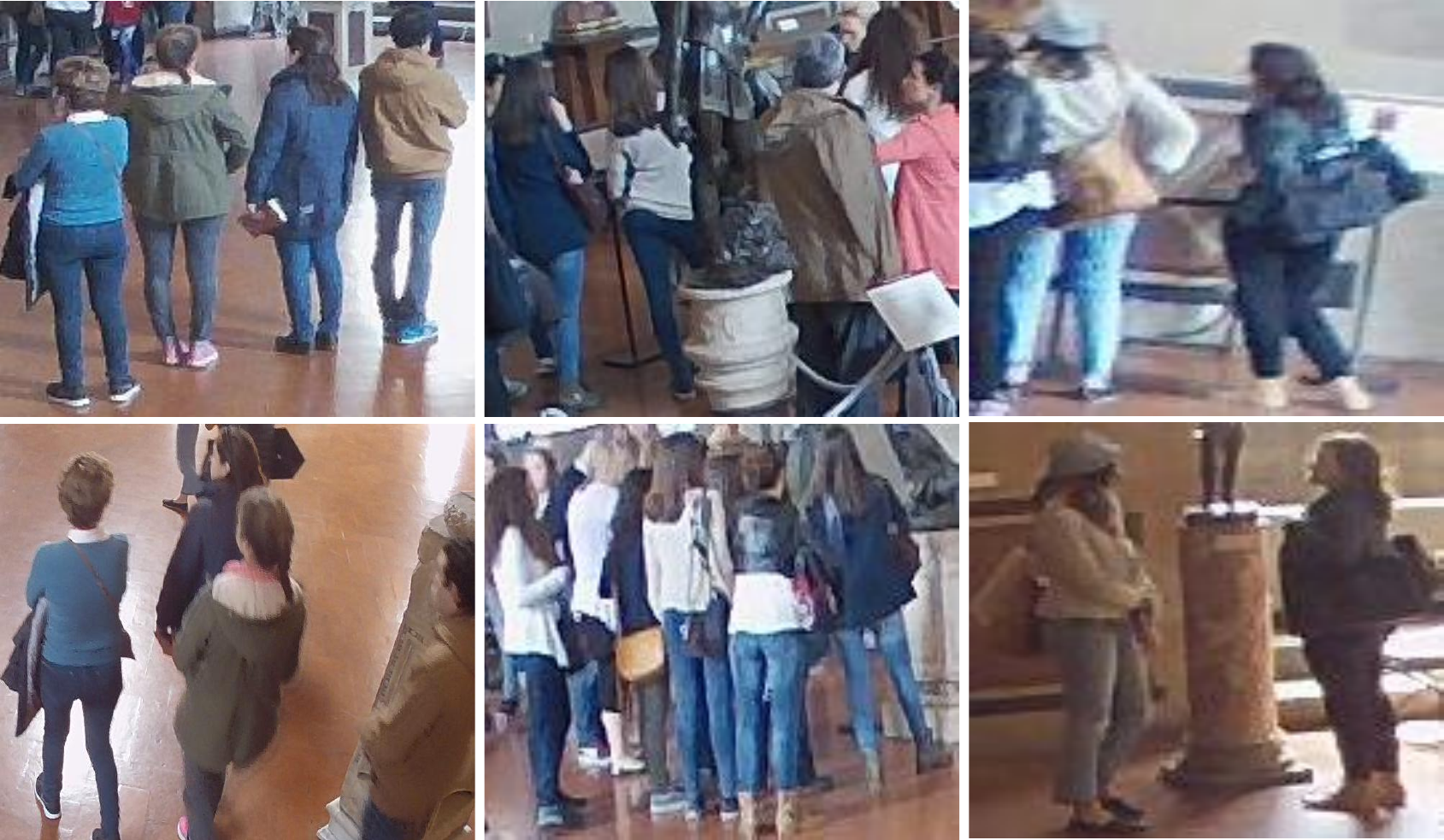}} \hfill
        \subfigure[OGRE]{\label{fig:ogri_samples}\includegraphics[width=0.325\linewidth]{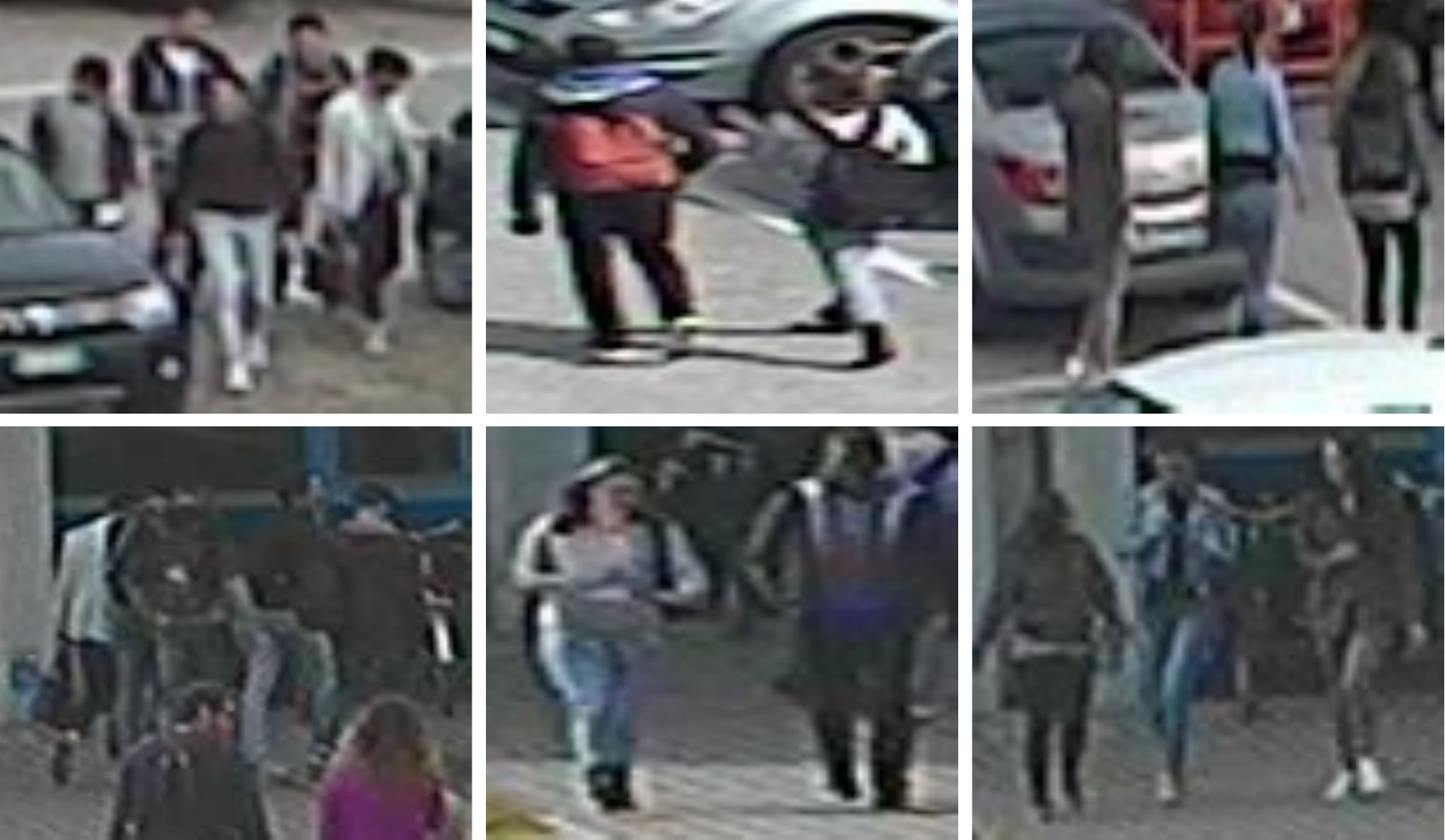}}
	\caption{Image samples from the~\subref{fig:ilids_samples} i-LIDS groups,~\subref{fig:museum_samples} Museum groups, and~\subref{fig:ogri_samples} OGRE datasets. Each column represents a same group, while each row depicts the image acquired by a different camera.}
\end{figure*}

\section{Experimental Results}
In this section we report on a series of experiments to assess the performance of the proposed method. From now on we refer to our solution as: Pooling Residuals of Encoded Features (PREF).

Plenty of single-person re-identification datasets have been publicly released --each one with different characteristics-- but just one of them is for group re-identification, namely the i-LIDS groups dataset~\cite{Zheng2009a}.
In order to evaluate the proposed solution under different scenarios, we collected two additional group datasets.

\vspace{1pt}
\noindent\textbf{i-LIDS Groups Dataset:}
This dataset has been obtained from the i-LIDS MCTS dataset which was captured at an airport arrival hall in the busy times under a multi-camera CCTV network.
The authors of~\cite{Zheng2009a} extracted 274 images of 64 groups.
Most of the groups have 4 images, either from different camera views or from the same camera but captured at different locations at different times.
Sample images for this dataset are shown in~\figurename~\ref{fig:ilids_samples}.

\vspace{1pt}
\noindent\textbf{Museum Groups Dataset\footnote{https://github.com/glisanti}:}
This dataset has been acquired in the hall of a national museum through four cameras, with small or no overlap. The cameras are installed so as to observe the artworks present in the hall and capture groups during their visits.
The dataset contains 524 manually annotated images of 18 groups, composed by a variable number of persons. Each group has about 30 images distributed between each one of the four cameras. Some samples are shown in~\figurename~\ref{fig:museum_samples}. 

\vspace{1pt}
\noindent\textbf{Outdoor Groups Re-Identification Dataset (OGRE)\footnote{https://github.com/iN1k1}:}
This dataset contains images of 39 groups acquired by three disjoint cameras pointing at a parking lot.
This results in approximatively 2,500 images acquired at different time instants and with different weather conditions.
The dataset has been acquired through a weakly supervised approach in which, given a manually selected group region, subsequent detections are obtained by running the KCF tracker~\cite{Henriques2015}. This results in a set of coarsely segmented group images that better resemble a real world scenario. 
Moreover, the dataset has severe viewpoint changes and a large number of self-occlusions (see~\figurename~\ref{fig:ogri_samples} for few samples).  

\subsection{Evaluation Protocol and Settings}
\noindent\textbf{Protocol:}
Tests are conducted following a single-vs-single shot scheme: for each group, one randomly selected image is included in the gallery, all the remaining images form the probe set.
As commonly performed~\cite{Bazzani2012,Zheng2015c,Liao2015}, such a process is repeated 10 times, then average results are computed.

\vspace{1pt}
\noindent\textbf{Performance Measure:}
All the results are reported in terms of Cumulative Matching Characteristic (CMC) curves and normalized Area Under Curve (nAUC) values. The CMC curve represents the expectation of finding the correct match in the first $r$ matches, whereas the nAUC gives a comprehensive measure on how well a method performs independently from the considered dataset.

\vspace{1pt}
\noindent\textbf{Source Datasets:} Three of the most commonly used single person re-identification datasets are employed as source domain from which to learn the dictionary of visual words.
Among all the possible ones, we selected the
i) ETHZ~\cite{Schwartz2009a} dataset since it contains multiple images of a same person from a similar viewpoint,
ii) CAVIAR~\cite{CAVIAR2004} dataset because of the low-resolution and occluded multiple images of a same person, and
iii) VIPeR~\cite{Gray2007a} dataset due to its challenging pose and illumination variations. 

\vspace{1pt}
\noindent\textbf{Dictionary Learning:}
We set $\lambda = 0.1$ because we did not notice significant changes in the performance with other values, while for the number of atoms, we run different experiments with $k \in \{300, 500, 1000\}$ when comparing with state-of-the-art in Sec.~\ref{sec:expsota}.

\begin{table*}[t]
\footnotesize
\caption{Rank-1 accuracy for different training datasets (\textbf{E} = ETHZ, \textbf{V} = VIPeR, \textbf{C} = CAVIAR), encoding distances and pooling of the residuals. Results are obtained on the i-LIDS groups dataset using PREF with 500 atoms and 50 PCA components. In parenthesis the nAUC. Best performances are marked with bold, whereas the second bests are marked with underline. Values are in percentage.\label{tab:all}}
\begin{tabulary}{1\linewidth}{L{4.3em}|C{5.5em}|C{5.5em}|C{5.5em}|C{5.5em}|C{5.5em}|C{5.5em}|C{5.5em}|C{5.5em}}
\toprule
& \multicolumn{2}{c|}{\textbf{L1}} & \multicolumn{2}{c|}{\textbf{Cosine}} & \multicolumn{2}{c|}{\textbf{Chi Square}} & \multicolumn{2}{c}{\textbf{Euclidean}}\\
\midrule
\emph{Dataset} & \emph{Max} & \emph{Average} & \emph{Max} & \emph{Average} & \emph{Max} & \emph{Average} & \emph{Max} & \emph{Average}  \\
\hline
\textbf{E} & 18.1 (73.3)  & 28.4 (77.9)  & 24.7 (75.8)  & 31.2 (77.8)  & 17.2 (72.7)  & 28.1 (77.9)  & 14.5 (72.1)  & 28.5 (77.8) \\ \hline
\textbf{V} & 18.4 (72.8)  & 29.7 (78.3)  & 21.6 (75.7)  & 29.5 (78.2)  & 17.1 (71.6)  & 28.5 (77.7)  & 13.7 (70.6)  & 27.2 (77.9) \\ \hline
\textbf{C} & 14.5 (71.7)  & 27.6 (77.2)  & 19.7 (76.8)  & 29.6 (77.8)  & 13.8 (69.9)  & 25.5 (77.4)  & 12.2 (68.8)  & 26.5 (77.1) \\ \hline
\textbf{E + V} & 18.4 (72.5)  & 28.5 (77.8)  & 23.7 (75.7)  & 30.5 (78.0)  & 16.6 (71.0)  & 29.5 (78.0)  & 13.9 (70.8)  & 27.2 (77.3) \\ \hline
\textbf{E + C} & 15.4 (72.5)  & 29.1 (78.3)  & 22.3 (75.5)  & 30.6 (78.1)  & 15.7 (71.6)  & 28.8 (77.4)  & 14.3 (68.9)  & 27.8 (77.4) \\ \hline
\textbf{V + C} & 17.3 (71.9)  & 28.1 (\textbf{78.7})  & 25.4 (77.0)  & \textbf{31.4} (78.1)  & 15.0 (71.5)  & 28.9 (78.5)  & 13.8 (71.2)  & 27.4 (\underline{78.6}) \\ \hline
\textbf{E + V + C} & 18.2 (72.9)  & 30.1 (78.4)  & 24.6 (77.1)  & \underline{31.1} (\textbf{78.7})  & 18.3 (73.2)  & 29.5 (77.9)  & 17.1 (71.3)  & 28.5 (78.1) \\
\bottomrule
\end{tabulary}
\end{table*}

\subsection{Ablation Study}
In this section, we thoroughly show how the performance of the proposed approach vary depending on the source dataset(s) considered for training, the distances used for residuals computation, the pooling method and the number of PCA components.
The analysis is carried out considering the i-LIDS groups dataset.
To run all the following experiments, we considered $k=500$. 

\vspace{1pt}
\noindent\textbf{Source Datasets and Residuals:}
To evaluate the performance of our solution considering different combinations of single person datasets, and different distances and pooling functions, we have computed the results in Table~\ref{tab:all}.

Results demonstrate that by considering more source datasets the overall performances tend to improve.
This might indicate that more discriminative atoms can be learned by considering  heterogeneous visual patches together with a robust sparse reconstruction.

As regards distances and pooling, the best results are obtained if average pooling is considered along with the cosine distance.
Such an outcome should be attributed to the fact that average pooling is able to better handle noisy assignments.
Similarly, the cosine distance is suitable because of the nature of the learned dictionary atoms~\cite{efron2004least}.

\begin{table}[t]
\centering
\footnotesize
\caption{Re-Identification using different number of PCA components. Results are obtained on i-LIDS groups using PREF with 500 atoms, ETHZ, VIPeR and CAVIAR for training and average pooling of cosine residuals. Best performance is marked with bold, the second best is underlined. Values are in percentage.\label{tab:pca}}
\begin{tabulary}{1\linewidth}{C{5.4em}|C{3.8em}|C{3.8em}|C{3.8em}||C{3.2em}}	
\toprule
\textbf{Components} & \textbf{Rank-1} & \textbf{Rank-10} & \textbf{Rank-25} & \textbf{nAUC} \\\midrule
20&29.4&58.5&75.1&78.2\\\hline
30&30.1&\underline{59.9}&\textbf{76.3}&\textbf{78.8}\\\hline
40&30.2&59.6&75.6&78.6\\\hline
50&\textbf{31.1}&\textbf{60.3}&75.5&\underline{78.7}\\\hline
60&\underline{30.7}&\textbf{60.3}&\underline{76.0}&\underline{78.7}\\\bottomrule
\end{tabulary}
\end{table}

\vspace{1pt}
\noindent\textbf{PCA Components:}
In Table~\ref{tab:pca} the performance of our solution are evaluated with varying number of PCA components.
Results show that the overall performances improve little when increasing the value of such an hyperparameter (\ie, there is an nAUC improvement of $0.5$ only).
Similar results are shown if the rank-1 indicator is considered.
This demonstrates that our solution does not hinge on the selection of such a value and is robust to the many group re-identification challenges even if only $20$ principal components are considered.

\begin{figure}
	\centering
	\subfigure[]{\label{fig:ilids_feat}\includegraphics[width=0.495\linewidth]{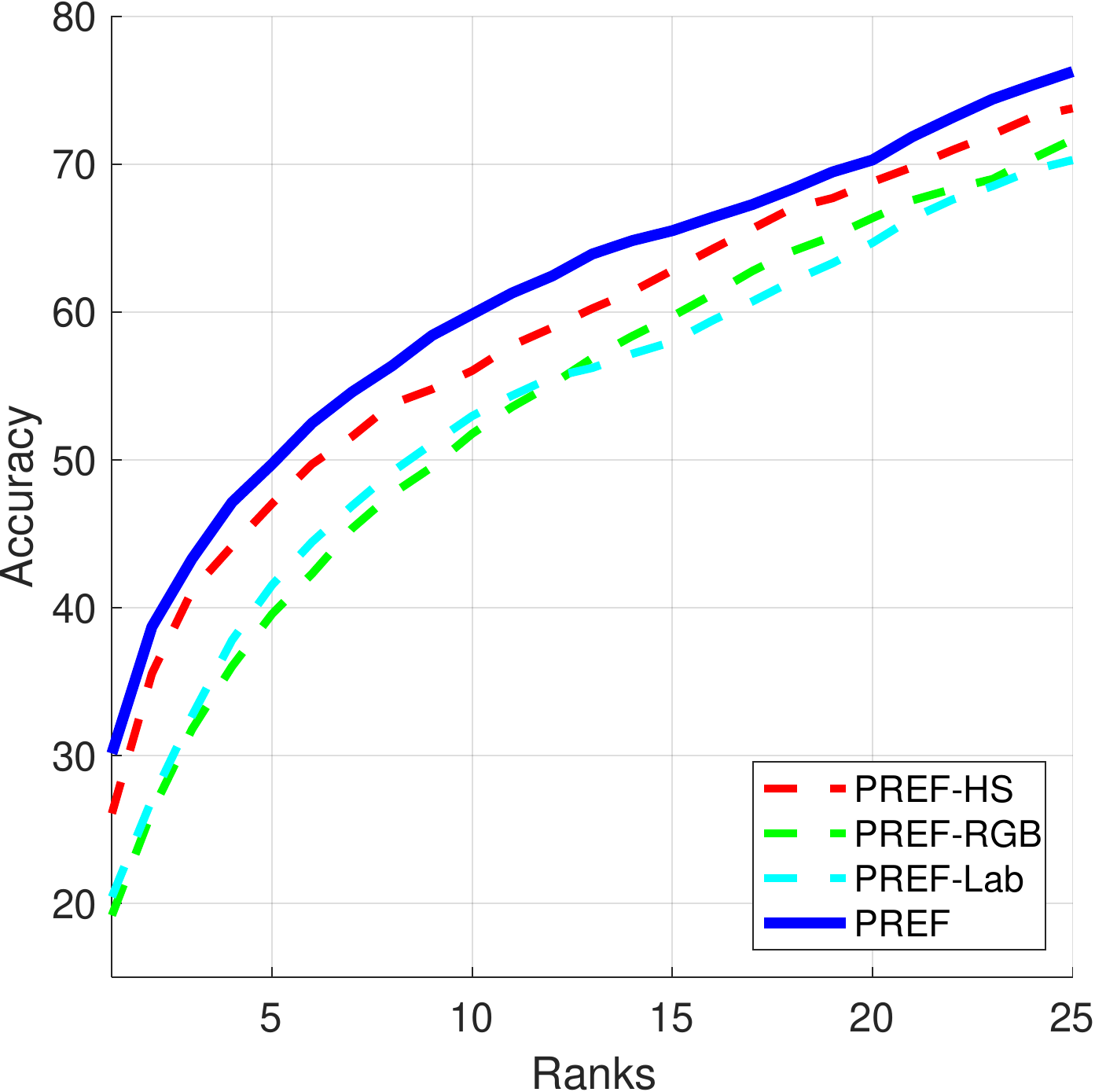}} \hfill
		        \subfigure[]{\label{fig:ilids_sota}\includegraphics[width=0.495\linewidth]{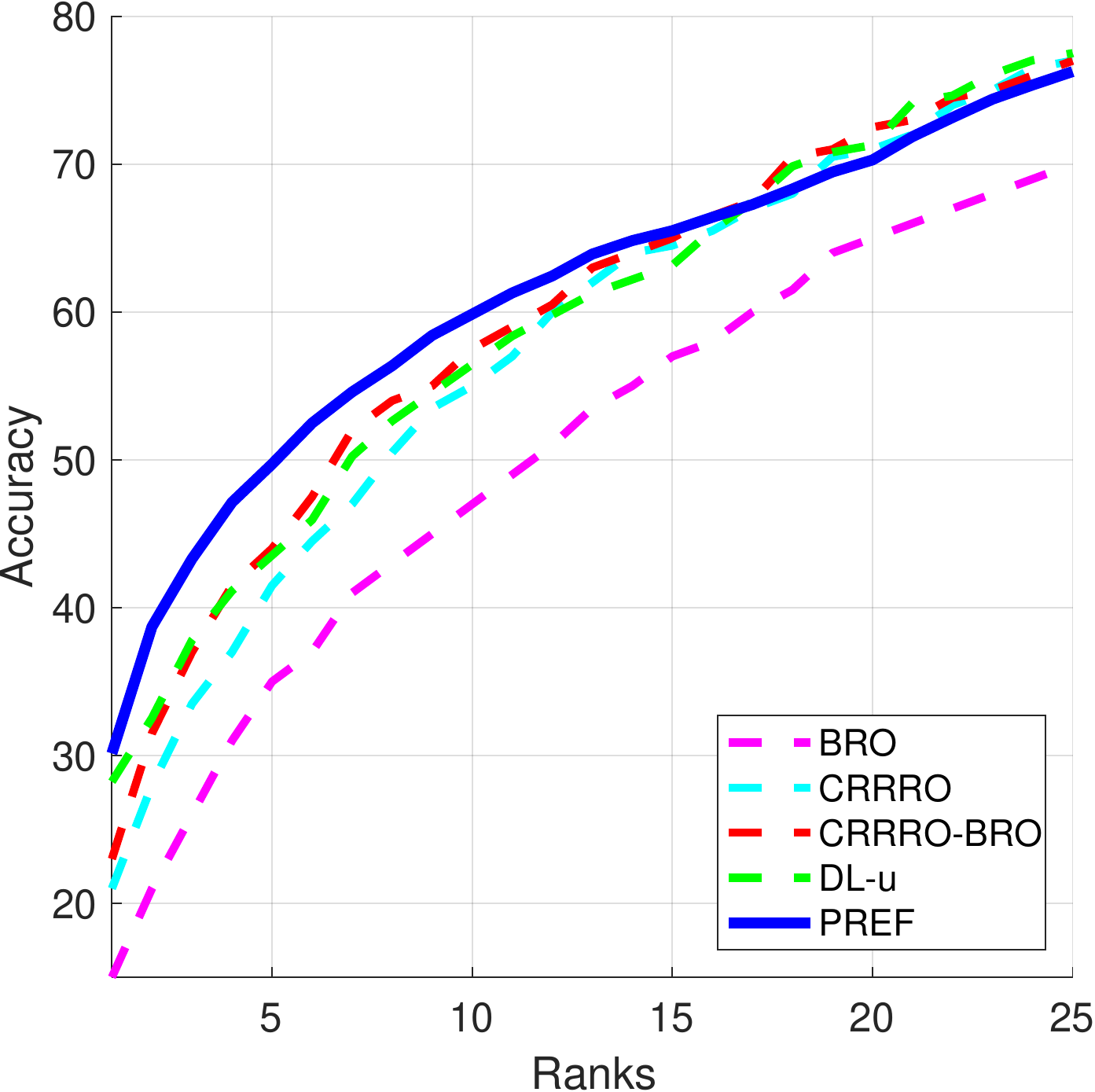}}
	\caption{CMC curves for the i-LIDS groups~\subref{fig:ilids_feat} obtained using the encoded features for each color histogram and their fusion; and~\subref{fig:ilids_sota} in comparison with state-of-the-art.}
\end{figure}

\vspace{1pt}
\noindent\textbf{Features:}
In \figurename~\ref{fig:ilids_feat}, we show the contribution of each encoded color histogram feature and their combination.
It is possible to appreciate that considering features projected onto the HS color space yields to the best results both in terms of rank-1 as well as nAUC.
However, as demonstrated by the literature~\cite{Xiong2014,Paisitkriangkrai2015}, considering all the color spaces helps in improving the overall performance.

\begin{figure*}
	\centering
	\includegraphics[width=0.94\textwidth]{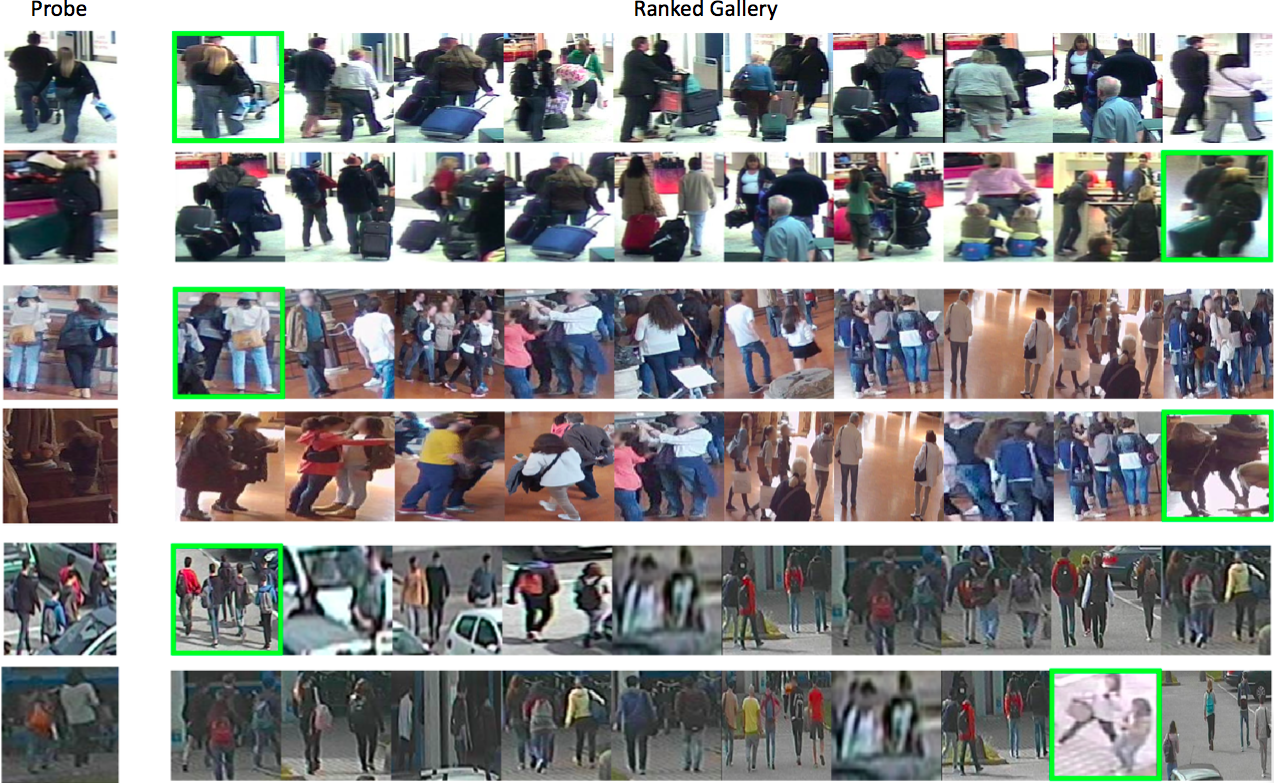}
\caption{Qualitative samples from the i-LIDS groups (\emph{top rows}), Museum groups (\emph{middle rows}) and OGRE datasets (\emph{bottom rows}). The correct match is highlighted in green and ranked galleries are sorted from left (Rank-1) to right (Rank-10). \label{fig:qualitative} }
\end{figure*}

\vspace{1pt}
\noindent\textbf{Qualitative Performance:}
Qualitative samples, showing the ranked gallery groups for critical probe images are reported in figures~\ref{fig:qualitative} for the i-LIDS groups, Museum groups and OGRE datasets, respectively.
Results show that our solution is able to handle situations in which subjects are exchanging their relative displacement as well as cases with severe occlusions.
However, drastic illumination variations challenge our approach since direct feature matching is not strong enough to tackle the feature transformation between cameras.
Such an issue could be addressed by exploiting metric learning solutions~\cite{Kostinger2012,Liao2015,Zheng2016,Martinel2015}.

\subsection{Comparison with State-of-the-Art}
\label{sec:expsota}
In the following, we report on the comparison with state-of-the-art in group re-identification and feature encoding.
\begin{table*}[t]
\small
\scriptsize
\caption{Re-Identification results on i-LIDS groups, Museum groups and OGRE  datasets. Results are obtained using ETHZ, VIPeR and CAVIAR for training, average pooling of cosine residuals and 50 components for PCA. \textbf{C} = number of clusters used for the encoding; \textbf{A} = number of atoms used for the dictionary learning. Best performance in bold, the second best is underlined. Values are in percentage.\label{tab:baseline}}
\begin{tabulary}{1\linewidth}{l|c|c|c|c|c|c|c|c|c|c|c|c|c}
\toprule
& &\multicolumn{4}{c|}{i-LIDS groups} & \multicolumn{4}{c|}{Museum groups} & \multicolumn{4}{c}{OGRE}\\
\midrule
\textbf{Method} & \textbf{C/A} & \textbf{Rank-1} & \textbf{Rank-10} & \textbf{Rank-25} & \textbf{nAUC} & \textbf{Rank-1} & \textbf{Rank-5} & \textbf{Rank-15} & \textbf{nAUC} & \textbf{Rank-1} & \textbf{Rank-10} & \textbf{Rank-25} & \textbf{nAUC} \\
\hline
IFV~\cite{Sanchez2013}&64&26.3&58.6&74.4&77.2&24.1&\underline{50.0}&87.6&\underline{62.7}&\underline{14.6}&\textbf{43.3}&\textbf{76.8}&\textbf{62.4}\\\hline
IFV~\cite{Sanchez2013}&128&26.1&\underline{60.2}&\textbf{75.8}&\underline{77.8}&23.6&49.0&\underline{88.4}&\textbf{62.8}&14.4&\underline{42.7}&75.6&\underline{61.8}\\\hline
IFV~\cite{Sanchez2013}&256&26.7&57.4&\underline{75.7}&76.8&24.8&49.2&88.3&62.6&14.1&42.4&\underline{75.9}&\underline{61.8}\\\hline
VLAD~\cite{Jegou2010b}&300&23.8&55.4&74.2&76.0&22.2&47.4&87.2&61.6&13.0&41.1&74.3&60.5\\\hline
VLAD~\cite{Jegou2010b}&500&24.6&54.0&75.6&76.5&23.0&47.6&88.2&61.8&12.6&40.4&73.8&59.9\\\hline
VLAD~\cite{Jegou2010b}&1000&26.0&57.0&75.0&76.7&22.9&48.4&\textbf{88.7}&62.4&12.3&39.6&73.2&59.6\\\hline
PREF&300&29.3&58.2&73.0&77.5&\underline{25.6}&49.8&87.3&62.6&14.3&41.0&74.9&61.0\\\hline
PREF&500&\textbf{31.1}&\textbf{60.3}&75.5&\textbf{78.7}&\textbf{25.8}&\textbf{50.2}&87.6&\underline{62.7}&\textbf{15.1}&41.6&75.8&61.6\\\hline
PREF&1000&\underline{30.1}&57.8&74.5&76.9&24.5&49.7&88.0&62.3&12.9&40.3&74.8&60.4 \\ \bottomrule
\end{tabulary}
\end{table*}

\vspace{1pt}
\noindent\textbf{Re-Identification:}
In~\figurename~\ref{fig:ilids_sota}, we report on the comparison with the state-of-the-art on the i-LIDS groups dataset.
Results show that the proposed solution outperforms existing approaches by about 8\% at rank-1, whereas at higher ranks similar performance is achieved.
It is worth noticing that the two group descriptors proposed in~\cite{Zheng2009a},~\ie, the Center Rectangular Ring Ratio-Occurrence Descriptor (CRRRO) and the Block based Ratio-Occurrence Descriptor (BRO), exploit shape features in addition to color ones.
More importantly, they proposed to learn a visual representation considering images that are from the same i-LIDS dataset. 
This might indicate that our approach is able to learn a robust visual representation from a source domain that is different from the target one.

In~\figurename~\ref{fig:ilids_sota}, we also report the CMC curve obtained with the unsupervised solution proposed in~\cite{Kodirov2015} based on dictionary learning (DL-u). This experiment has been conducted using the same features as in our solution, so as to have a fair comparison. Lower performance for~\cite{Kodirov2015} can be motivated by the fact that it considers all the patches as a unique descriptor, thus it hinges on the spatial displacement of persons within the image. 

\vspace{1pt}
\noindent\textbf{Feature Encoding:}
To have a more thorough with respect to the state-of-the-art, we performed experiments considering two encoding techniques, namely IFV~\cite{Sanchez2013} and VLAD~\cite{Jegou2010b}, and our group representation.
Experiments are conducted on the i-LIDS groups dataset and on the two newly introduced datasets.
For IFV and VLAD, we considered $\{64, 128, 256\}$ and $\{300, 500, 1000\}$ number of clusters, respectively.
For these two methods and the proposed solution we obtained the encoding model using ETHZ, VIPeR and CAVIAR datasets.

Results in Table~\ref{tab:baseline} demonstrate that the proposed encoding scheme has better rank-1 performance than existing approaches on all datasets.
We hypothesize that this result is due to the fact that the clustering solutions exploited to obtain the encoding models for IFV and VLAD are more sensitive to outliers (\ie, noise), whereas dictionary learning with sparse coding helps in reducing this effect~\cite{Mairal2009}.

\begin{table}[t]
\centering
\footnotesize
\caption{Results on i-LIDS groups dataset obtained using the same configuration adopted for Table~\ref{tab:baseline} and spatial information. \textbf{C} = number of clusters used for the encoding; \textbf{A} = number of atoms used for the dictionary learning. Best performance in bold, the second best is underlined. Values are in percentage.\label{tab:spatial_encoding}}
\begin{tabulary}{1\linewidth}{l|c|c|c|c|c}
\toprule
\textbf{Method} & \textbf{C/A} & \textbf{Rank-1} & \textbf{Rank-10} & \textbf{Rank-25} & \textbf{nAUC} \\
\midrule
IFV~\cite{Sanchez2013}&64&22.6&51.2&71.1&73.6\\\hline
IFV~\cite{Sanchez2013}&128&\textbf{24.5}&\textbf{53.4}&\textbf{73.4}&\textbf{75.5}\\\hline
IFV~\cite{Sanchez2013}&256&\underline{23.3}&\underline{52.1}&\underline{72.0}&\underline{74.2}\\\hline
VLAD~\cite{Jegou2010b}&300&18.2&48.8&72.4&73.2\\\hline
VLAD~\cite{Jegou2010b}&500&17.2&46.1&70.2&72.3\\\hline
VLAD~\cite{Jegou2010b}&1000&17.1&49.2&70.3&72.1 \\\hline
PREF&300&20.1&48.5&66.9&71.6\\\hline
PREF&500&21.7&49.6&67.5&71.9 \\\hline
PREF&1000&21.1&48.5&66.9&71.4 \\ \bottomrule
\end{tabulary}
\end{table}

\vspace{1pt}
\noindent\textbf{Spatial Encoding:}
To verify whether the proposed solution is robust to the spatial appearance ambiguities of group images (\eg, distinguishing two groups of people with opposite appearance), we have conducted the following experiment: to each 64-D feature extracted from each patch (Sec.~\ref{sec:group_features}) we have concatenated its $(x,y)$ position, thus producing a 66-D vector. The considered $(x,y)$ position of the patch is calculated with respect to the detected person image size. This avoids the problem of having an absolute $(x,y)$ information that depends on the person location within the group image.
Results in Table~\ref{tab:spatial_encoding}, show that IFV/VLAD/PREF performances degrade by about $7\%$ if such spatial information is included in the feature vector.
This might indicate that, spatially constraining the patches of a person may limit re-identification performance due to appearance variations caused by pose changes and the different viewpoints from which a person can be observed.

\section{Conclusion}
In this paper we have proposed a solution for associating group of persons across different cameras. The proposed solution grounds on the idea of transferring knowledge from single person re-identification to groups, in an unsupervised way, exploiting sparse dictionary learning. The sparse dictionary is learned from classical single person re-identification images. Then a sparsity-driven residual along with a pooling strategy have been introduced to encode features coming from the group and to obtain the final representation. An extensive evaluation shows that the proposed solution achieves state-of-the-art performance on the three datasets for group re-identification. Moreover, results show that it is worth investigating the introduction of a learning scheme to better handle cross-view re-identification issues.

\end{document}